\documentclass{article}

\usepackage{arxiv}

\usepackage[utf8]{inputenc} 
\usepackage[T1]{fontenc}    
\usepackage{hyperref}       
\usepackage{url}            
\usepackage{booktabs}       
\usepackage{amsfonts}       
\usepackage{nicefrac}       
\usepackage{microtype}      
\usepackage{lipsum}		
\usepackage{graphicx}
\usepackage[numbers]{natbib}
\usepackage{doi}
\usepackage{amsmath}

\usepackage{subfigure}

\title{Text Classification: A Perspective of Deep Learning Methods}

\date{October 28, 2020}	

\author{ {\includegraphics[scale=0.06]{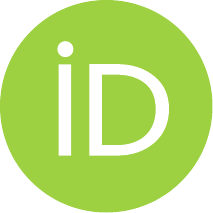}\hspace{1mm}Zhongwei Wan} \\
	Department of Aritificial Intelligence\\
   University of Chinese Academic of Science\\
	Beijing, China \\
	\texttt{11612201@mail.sustech.edu.cn} \\
}

\renewcommand{\shorttitle}{\textit{arXiv} Template}

\hypersetup{
pdftitle={Text Classification: A Review of Deep learning Methods},
pdfsubject={q-bio.NC, q-bio.QM},
pdfauthor={David S.~Hippocampus, Elias D.~Striatum},
pdfkeywords={First keyword, Second keyword, More},
}

\begin{document}
\maketitle

\begin{abstract}
 In recent years, with the rapid development of information on the Internet, the number of complex texts and documents has increased exponentially, which requires a deeper understanding of deep learning methods in order to accurately classify texts using deep learning techniques, and thus deep learning methods have become increasingly important in text classification. Text classification is a class of tasks that automatically classifies a set of documents into multiple predefined categories based on their content and subject matter. Thus, the main goal of text classification is to enable users to extract information from textual resources and process processes such as retrieval, classification, and machine learning techniques together in order to classify different categories. Many new techniques of deep learning have already achieved excellent results in natural language processing. The success of these learning algorithms relies on their ability to understand complex models and non-linear relationships in data. However, finding the right structure, architecture, and techniques for text classification is a challenge for researchers. This paper introduces deep learning-based text classification algorithms, including important steps required for text classification tasks such as feature extraction, feature reduction, and evaluation strategies and methods. At the end of the article, different deep learning text classification methods are compared and summarized.
\end{abstract}


\section{Introduction}
Text classification is a classical problem in natural language processing. The task is to assign predefined categories to a given sequence of texts. In recent years, the study of text classification has become increasingly important due to the rapid growth of social networks, blogs and forums, and the increase in the size of online academic libraries. As a result, text classification is widely used in information retrieval systems and search engine applications. At the same time, text classification can also be used for email and SMS spam filtering. Most of the text classification techniques include feature extraction of text, data reduction and deep learning model selection, and model evaluation. Also, text classification systems can classify text by its size, such as document level, paragraph level, sentence level, and clause level \citep{DBLP:journals/corr/abs-1904-08067}.

Before deep learning became the dominant model, traditional machine learning had a wide range of applications in text classification, such as using ensemble learning techniques like boosting and bagging for text classification and analysis \citep{DBLP:conf/icml/AbeM98}. At the same time, \citep{CHEN2017147} used simple logistic regression to classify textual information for information retrieval using simple logistic regression. \citep{DBLP:journals/jasis/Larson10} uses a Naive Bayesian classifier to classify documents because Naive Bayes uses less memory and computation, and is the classifier that is used more often by traditional machine learning.

Therefore, this review consists of several parts, the first part will briefly introduce several deep learning algorithms for feature extraction in text classification tasks, such as Word2Vec \citep{DBLP:journals/corr/abs-1301-3781}, Global Vectors for Word Representation (GloVe) \citep{DBLP:conf/emnlp/PenningtonSM14}, and several word embedding algorithms. In the second part, we will also briefly introduce several data reduction algorithms that may be used in traditional machine learning-based text classification tasks, such as Principal Component Analysis (PCA) \citep{Karamizadeh2009PrincipalCA}, Linear Discriminant Analysis (LDA) \citep{Hrault1984RseauDN}, which can improve the Accuracy in traditional machine learning text classification tasks, in some cases.We will focus on several conventional deep learning-based text classification algorithms, such as LSTM \citep{DBLP:journals/neco/HochreiterS97}, GRU \citep{DBLP:journals/corr/abs-1301-3781}, and several state of the art attention models, such as Transformer \citep{DBLP:conf/nips/VaswaniSPUJGKP17}, and improved versions based on Transformer such as XL-Net \citep{DBLP:journals/corr/abs-1906-08237}, Bert \citep{DBLP:journals/corr/abs-1810-04805}, and several improved models of Bert, among others. Finally, Several common model evaluation techniques are very essential in text classification tasks, such as accuracy, Fb score, receiver operating characteristics (ROC), and area under the ROC curve (AUC). 
\begin{figure}[h]
    \centering   
    \includegraphics[width=1\textwidth]{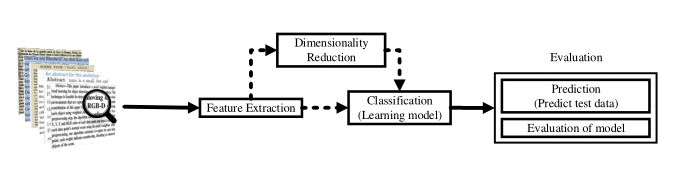}
    \caption{Pipeline of Text classification \citep{DBLP:journals/corr/abs-1904-08067}}
    \label{pipline of text classification}
\end{figure}

\section{Feature extraction}
 Word embedding is a key technique in the feature extraction process of text classification. Although we have used Tokenizer before the feature extraction task to divide sentences into words and count the number of occurrences of each word, as well as to generate syntactic word representations, this process does not capture the semantic information between words and words. This problem can cause serious problems in the model's understanding of the semantic information in sentences. For example, the n-gram model does not find word-to-word similarities. So google researchers in the journal NIPS solved this problem by word vector embedding. \citep{DBLP:journals/corr/abs-1301-3781} is one of the foundational papers of word2vec, presented by Tomas Mikolov of Google. The paper proposed two word2vec model structures, CBOW and Skip-gram. And \citep{DBLP:conf/nips/MikolovSCCD13} is another foundational paper on word2vec. The Skip-gram model is described in detail, including the specific form of the model and two feasible training methods, Hierarchical Softmax and Negative Sampling. \citep{DBLP:conf/emnlp/PenningtonSM14} presented GloVe, an improved technique of word2vec, using global matrix decomposition (LSA) and local content windows (Word2vec) to fully utilize statistical information to train the model using elements with non-zero frequencies in the word co-occurrence matrix.

From the three papers on word embedding presented by the above researchers, word embedding is a feature learning technique in which each word from a vocabulary is mapped to an X-dimensional vector. Two of the key techniques, word2vec and GloVe, have been successfully used in deep learning techniques. In word2vec, the training goal of the Skip-gram model is to find word representations that are useful for predicting words in the context of a sentence or document. And the formula is:
\begin{equation}
	\frac{1}{T} \sum_{t=1}^{T} \sum_{-c \leq j \leq c_{j} \neq 0} \log p\left(w_{t+j} \mid w_{t}\right)
\end{equation}
 The CBOW model uses the contextual words to predict the central words, and the diagram of the two models is shown below.
\begin{figure}[h]
    \centering   
    \includegraphics[width=0.5\textwidth]{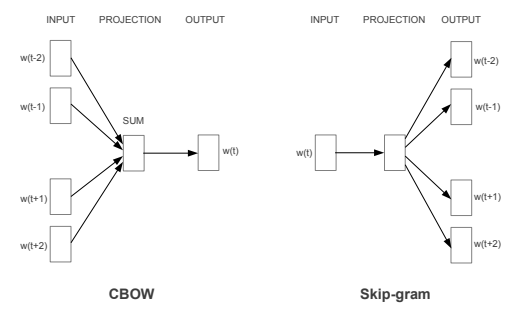}
    \caption{The CBOW architecture predicts the current word based on the
context, and the Skip-gram predicts surrounding words given the current word[5].}
    \label{wolf}
\end{figure}

\begin{figure}[h]
    \centering   
    \includegraphics[width=0.5\textwidth]{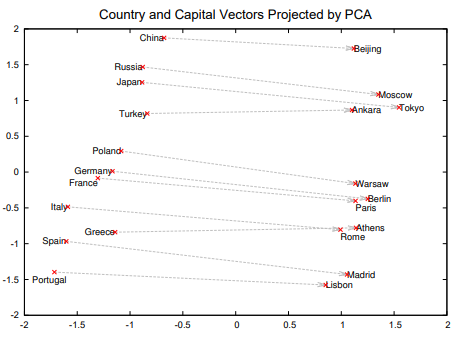}
    \caption{Two-dimensional PCA projection of the 1000-dimensional Skip-gram vectors of countries and their
capital cities[62]}
    \label{wolf}
\end{figure}

\section{Feature reduction}
 In this section, we will briefly introduce possible feature reduction techniques that can be used in text classification tasks. Many text sequences in term-based vector models consist of many complex features, and therefore many researchers use machine learning-based feature reduction techniques to reduce the feature space size and thus the temporal and spatial complexity of the model. We will briefly introduce two common dimensionality reduction techniques such as PCA [7], LDA [8] in the following.

\subsection{Principal Component Analysis (PCA)}
 The core idea of PCA is to reduce dimensionality by finding approximate subspaces of the data distribution. n-dimensional features are mapped by PCA to the k-dimension, which is a new orthogonal feature, also known as a principal component, that is reconstructed from the original n-dimensional features. k-dimensional features are reconstructed from the original n-dimensional features. It is closely related to the data itself. The first new axis is chosen to be the direction with the greatest variance in the original data, the second new axis is chosen to be the one with the greatest variance in the plane orthogonal to the first axis, and the third axis is the one with the greatest variance in the plane orthogonal to the 1st and 2nd axes. By analogy, n such axes can be obtained. The new axes are obtained in this way. Finally, by eigenvalue decomposition or SVD decomposition of the covariance matrix of the data, the eigenvector corresponding to the first K largest eigenvalues of the desired dimension is obtained, and multiplied with the original data matrix to obtain the dimensionality-reduced features.
\begin{equation}
Cov(X, Y)=E[(X-E(X))(Y-E(Y))]=\frac{1}{n-1} \sum_{i=1}^n\left(x_i-\bar{x}\right)\left(y_i-\bar{y}\right)
\end{equation}
\begin{equation}
A=U \Sigma V^{T}
\end{equation}

\subsection{Linear Discriminant Analysis (LDA)}
 LDA differs from the unsupervised learning technique of PCA in that it is a supervised dimensionality reduction technique, which means that each sample of the dataset has a category output.The core idea of LDA is that we want to project the data in a low dimension, and after projection we want the projection points of each category of data to be as close to each other as possible, and the distance between the category centers of the different categories of data to be as large as possible. Since we are projecting multiple categories to a low dimension, the projected low dimensional space is not a straight line, but a hyperplane. Suppose we project a low-dimensional space of dimension d, corresponding to a base vector of (w1,w2,...wn). The core formula of LDA is as follows, Sw is defined as a class scatter matrix and Sb is defined as an interclass scatter matrix, so the optimization function of the LDA dimensionality reduction algorithm is as follows.
\begin{equation}
\underbrace{\arg \max }_W J(W)=\frac{\prod_{\text {diag }} W^T S_b W}{\prod_{\text {diag }} W^T S_w W}
\end{equation}
\begin{equation}
S_{b}=\sum_{j=1}^{k} N_{j}\left(\mu_{j}-\mu\right)\left(\mu_{j}-\mu\right)^{T}
\end{equation}
\begin{equation}
S_{w}=\sum_{j=1}^{k} S_{w j}=\sum_{j=1}^{k} \sum_{\mathbf{k} \in X_{j}}\left(x-\mu_{j}\right)(x-\mu j)^{T}  
\end{equation}

\section{Deep Learning models}
 Deep learning models have achieved state-of-the-art results in many fields and are in many ways superior to traditional machine learning algorithms. Traditional machine learning algorithms require a feature extraction and classifier selection process, which increases the cost of labor. Deep learning models are end-to-end training models for many computer vision and natural language tasks. In many tasks, deep learning models have much better fitting and generalization capabilities than traditional machine learning algorithms, and we will introduce the most common deep learning backbone models, LSTM \citep{DBLP:journals/neco/HochreiterS97}, GRU \citep{DBLP:journals/corr/abs-1301-3781}, and Transformer \citep{DBLP:conf/nips/VaswaniSPUJGKP17}, which are mainly used in text classification tasks. ELMO \citep{Peters:2018}, GPT \citep{Radford2018ImprovingLU}, bert \citep{DBLP:journals/corr/abs-1810-04805}, GPT2 \citep{radford2019language}, XL-net \citep{DBLP:journals/corr/abs-1906-08237} and other state of the art deep learning models.

\subsection{LSTM and GRU}
Recurrent neural networks (RNNs) are the most commonly used neural network architectures for text data mining and classification, especially for serial data such as textual information. Thus, RNNs have advantages in classifying text, string, and sequential data, and in considering information from previous nodes on a temporal order basis.
\subsubsection{Long Short Time memory(LSTM)}
 LSTM is an improved model of RNN, which, compared to the original RNN, better overcomes the gradient disappearance problem through structures such as input gates, memory cells, forgetting gates, output gates, and hidden units, thereby maintaining long-term dependence. As shown in Figure 4, the forgetting gate controls whether the information from the cell memory of the previous time step is passed to the current time step, while the input gate controls how the input information of the current time step flows into the memory cell of the current time step through the candidate memory cell. This design can cope with the gradient decay problem in cyclic neural networks and better capture the dependency of large time step distances in time series. Therefore, the basic formula for the LSTM model is as follows.
\begin{figure}[h]
    \centering   
    \includegraphics[width=0.3\textwidth]{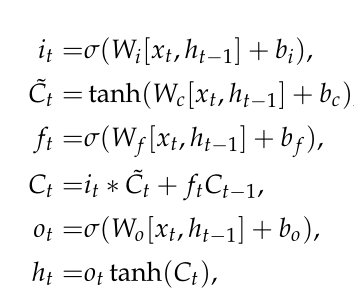}
    \label{pipline of text classification}
\end{figure}
\subsubsection{Gate Recurrent Unit(GRU)}
 Compared to LSTM, GRU is a simplified variant of LSTM with a reset gate and an update gate. As shown in figure 4, the state of the previous time step is discarded when the element in the reset gate is close to 0, and the hidden state of the previous time step is retained when the element in the reset gate is close to 1. The update gate controls how the hidden state should be updated by the candidate hidden state that contains information about the current time step. Finally, the hidden state of the current time step is a combination of the update gate of the current time step and the hidden state of the previous time step. The structure of GRU is as follows.

\begin{figure}[htbp]
\centering
\subfigure{
\includegraphics[width=5.5cm]{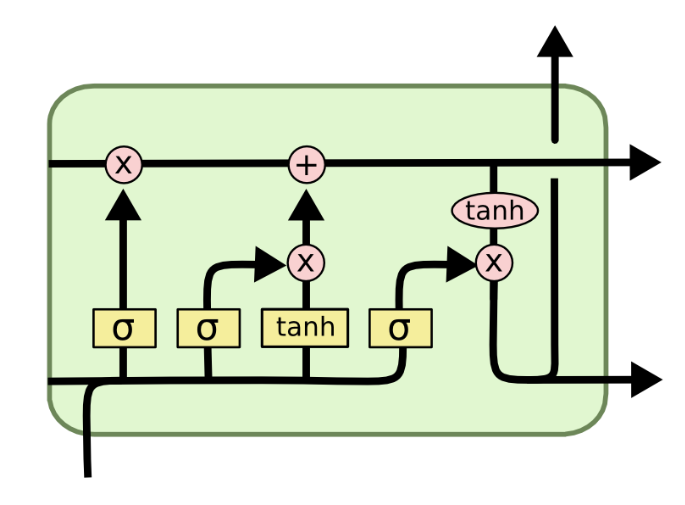}
}
\quad
\subfigure{
\includegraphics[width=7.2cm]{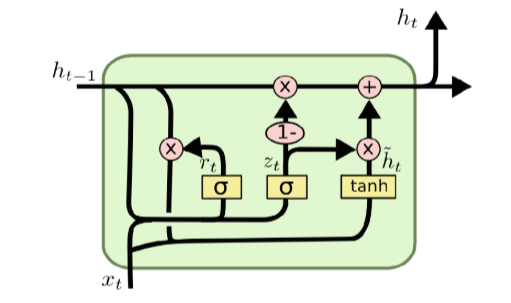}
}
\caption{The cell of LSTM and GRU}
\end{figure}


\subsection{ELMo}
 ELMo is a pre-training model based on Bi-direction LSTM. In previous word2vec work, each word corresponds to only one word vector, but it is not useful for polysemous words with complex semantic information. Therefore, in ELMo, the pre-trained model is no longer just a correspondence of vectors. When using ELMo, a sentence or a paragraph is entered into the model, and the model inferred the word vector for each word based on the context. The formula for the ELMo model is as follows.
\begin{equation}
    p\left(t_{1}, t_{2}, \ldots, t_{N}\right)=\prod_{k=1}^{N} p\left(t_{k} \mid t_{1}, t_{2}, \ldots, t_{k-1}\right)
\end{equation}

\begin{equation}
    p\left(t_{1}, t_{2}, \ldots, t_{N}\right)=\prod_{k=1}^{N} p\left(t_{k} \mid t_{k+1}, t_{k+2}, \ldots, t_{N}\right)
\end{equation}
\begin{equation}
    \begin{array}{l}
\sum_{k=1}^{N}\left(\log p\left(t_{k} \mid t_{1}, \ldots, t_{k-1} ; \Theta_{x}, \vec{\Theta}_{L S T M}, \Theta_{s}\right)\right. \\
\left.\quad+\log p\left(t_{k} \mid t_{k+1}, \ldots, t_{N} ; \Theta_{x}, \overleftarrow{\Theta}_{L S T M}, \Theta_{s}\right)\right)
\end{array}
\end{equation}

\begin{figure}[h]
    \centering   
    \includegraphics[width=0.6\textwidth]{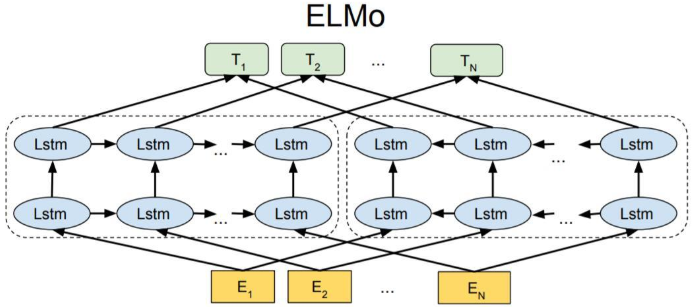}
    \caption{The model of ELMo}
    \label{pipline of text classification}
\end{figure}

 Equation 7 computes the objective function to be learned by the forward LSTM language model, while Equation 8 computes the objective function for the backward LSTM language model. The Bi-directional LSTM language model used by ELMo combines the forward and backward formulas to maximize the forward and backward maximum likelihood probabilities, as shown in Equation 9. Thus, ELMo's model, shown in Fig. 9, has two improvements on the original LSTM, the first being the use of a multilayer LSTM and the second being the addition of a backward language model. For the backward language model, it is similar to the forward language model in that the backward context is given to predict the forward context. When ELMo's model completes the pre-training task, it can be used for other NLP tasks, or the Bi-directional LSTM can be fine-tuned and the representation of the task can be improved. This is a transfer learning method that can be used for NLP text classification tasks.

\subsection{Transformer}
 Attention is all you need. The Transformer model architecture, which has recently gained great interest among researchers, is based on a total attention model and has validity in the areas of language, vision, and reinforcement learning. Particularly in the field of natural language, the Transformer model has gone beyond RNN and has taken state of the art effects in several NLP tasks. The most important part of the Transformer model is the self-attention mechanism, which can be viewed as a graph-like induction bias that connects all the markers in a sequence through association-based pooling operations. Based on self-attention, the researchers proposed a multiheaded attention mechanism, a position-wise feed-forward network, layer normalization modules and residual connectors. Input to the Transformer model is often a tensor of shape (batch size, sequence length). The input first passes through an embedding layer that converts each one-hot token representation into a d dimensional embedding, And than an positional encodings is added to the aforementioned new tensor. The formula for the Transformer is as follows.

    

\begin{equation}
    X_{A}=\text { LayerNorm(MultiheadSelfAttention } \left(X)\right)+X
\end{equation}

\begin{equation}
    X_{B}=\text { Layer Norm(PositionFFN } \left.\left(X_{A}\right)\right)+X_{A}
\end{equation}

\begin{equation}
    \text { Attention }(Q, K, V)=\operatorname{softmax}\left(\frac{Q K^{T}}{\sqrt{d_{k}}}\right) V
\end{equation}

 where, Q, K, V are linear transformations applied on the
the temporal dimension of the input sequence. dk is the dimension of the vector in the multiheaded attention model and is multiplied by 1/$\sqrt{dk}$ to counteract the fact that when dk is large, the size of the dot product increases, thus pushing the softmax function into the region of its very small gradient. Thus, Transformer achieves the effect of state of the art on the sequence to sequence translation task. At the same time, for the text classification task, the use of transformer's encoder model coupled with the multilayer perceptron network (MLP) for the classification regression task also achieves good results.
\begin{figure}[h]
    \centering   
    \includegraphics[width=0.6\textwidth]{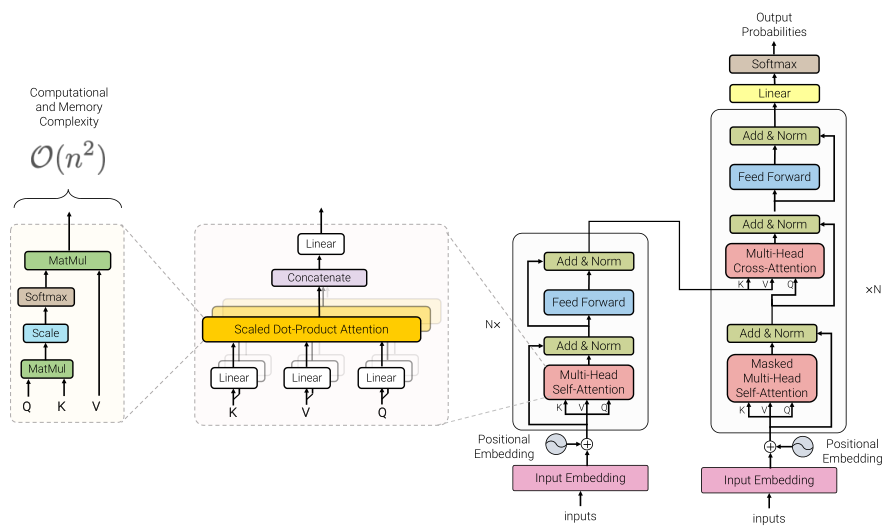}
    \caption{The model of Transformer}
    \label{pipline of text classification}
\end{figure}

\subsection{Generative Pre-Training (GPT)}
 Open AI proposes the Generative Pre-Training (GPT) model, a language comprehension task that uses a semi-supervised approach to processing. The GPT task is to learn a generic language representation, which can then be fine-tuned for use in many downstream tasks, such as natural language processing tasks like text classification. The approach to unsupervised text processing is to maximize the great likelihood of the language model, hence the Transformer's decoder language model is used in the paper. Unlike Transformer's encoder, the Mask multiple-head attention model in the decoder model uses a mask that allows the model to notice only previous sequences during pre-training. Thus, the multi-layer structure of GPT applies multi-headed self-attention to process the input text plus a feedforward network of location information, and the output is a conceptual distribution of words.

Since the GPT uses a one-directional Transformer model, the model can only see the words above. The training process is to add Positional Encoding to the N-word word vector of the sentence and input it into the Transformer mentioned above, and the N outputs predict the next word at that location. The formula for the model is as follows.
\begin{equation}
    L_{1}(\mathcal{U})=\sum \log P\left(u_{i} \mid u_{i-k}, \ldots, u_{i-1} ; \Theta\right)
\end{equation}
\begin{equation}
h_{0} =U W_{e}+W_{p}
\end{equation}
\begin{equation}
h_{l} =\text { transformerblock }\left(h_{l-1}\right) \forall i \in[1, n]
\end{equation}
\begin{equation}
P(u) =\operatorname{Softmax}\left(h_{n} W_{e}^{T}\right)
\end{equation}

\begin{figure}[htbp]
\centering
\subfigure{
\includegraphics[width=2.5cm]{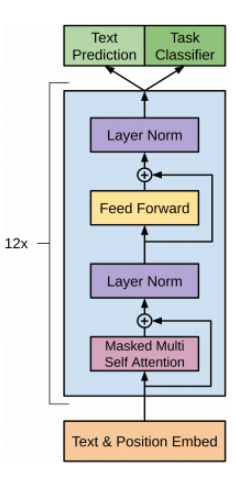}
}
\quad
\subfigure{
\includegraphics[width=5.5cm]{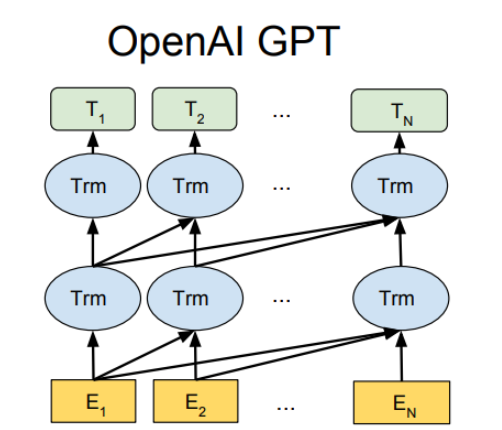}
}
\caption{The model of GPT}
\end{figure}

\subsection{Bi-directional Encoder Representation from Transformer(Bert)}
 BERT is very similar to GPT in that it is a two-stage Transformer-based training model, divided into Pre-Training and Fine-Tuning stages. The parameters in this model are fine-tuned to adapt it to different downstream tasks. However, GPT uses a unidirectional Transformer, while BERT uses a bidirectional Transformer, which means no Mask operation is required. In addition, BERT uses the Encoder in the Transformer model, while GPT uses the Decoder in the Transformer model, so the pre-training methods are different for the two models. In addition, BERT uses the Masked Language Model (MLM) pre-training method and the Next Sentence Prediction (NSP) pre-training method, which can be trained at the same time.

In order to distinguish between two sentences, BERT adds a Segment Embedding to be learned during pre-training, in addition to Positional Encoding. In this way, BERT's input consists of a word vector, a position vector, and a segment vector that are added together. In addition, the two sentences are distinguished from each other using <SEP> tags. The embedding is as follow.

\begin{figure}[h]
    \centering   
    \includegraphics[width=0.6\textwidth]{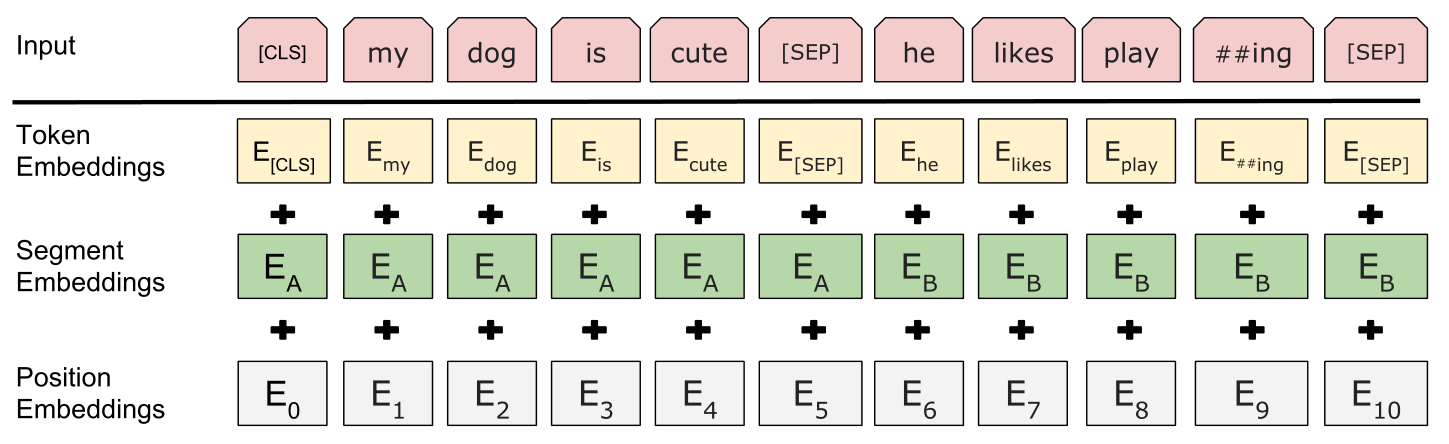}
    \caption{The Embedding of Bert}
    \label{pipline of text classification}
\end{figure}

BERT's Fine-Tuning phase is not much different from GPT. Because of the bi-directional Transformer, the auxiliary training target used by GPT in the Fine-Tuning phase, i.e., the language model, has been abandoned. In addition, the output vector for classification prediction has been changed from the output position of the last word in GPT to the position of <CLS> at the beginning of a sentence.

\begin{figure}[h]
    \centering   
    \includegraphics[width=0.6\textwidth]{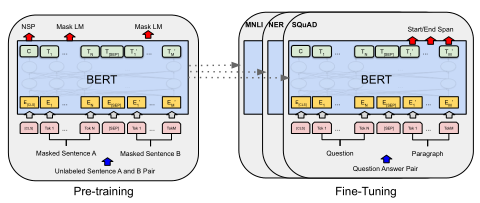}
    \caption{The fine-tune of Bert}
    \label{pipline of text classification}
\end{figure}

\subsection{XL-Net}
XLNet combines two pre-trained model ideas, Bert and GPT, to present a state-of-the art deep learning model in the field of natural language processing. From the above, we can see that GPT is a typical autoregressive language model, which has the disadvantage of not being able to use the above and below information at the same time. Bert, on the other hand, belongs to the Autoencoder Language Model, where Bert randomly Masks out a portion of the words in the input sequence, and then one of the main tasks of the pre-training process is to predict the Masks based on the contextual words. Therefore, XLNet needs to improve on Bert because the first pre-training stage takes the training mode of introducing [Mask] markers to Mask off some words, while the fine-tuning stage does not see such forced Mask markers, so the two stages have inconsistent usage patterns, which may lead to some performance loss. Another is that Bert assumes in the first pre-training phase that multiple words are Masked out of the sentence, that there is no relationship between the Masked out words, that they are conditionally independent, and that sometimes there is a relationship between the words, which XLNet takes into account.

Based on the autoregressive model, XLNet introduces the training objective of the Permutation Language Model in the pre-training phase, assuming that the sequence is x1,x2,x3,x4, the word to be predicted is x3, and ContextBefore is x1,x2. To take into account the content of ContextAfter, an After the random permutation operation, the sequences x4,x2,x3,x1 are obtained and the model is input. In this operation, XLNet is able to take into account both the context and the content. This part of the improvement is achieved through the mask operation in Transformer Attention. We refer to the literature for the detailed implementation process. The main idea of mask attention is as follow:
\begin{figure}[h]
    \centering   
    \includegraphics[width=0.6\textwidth]{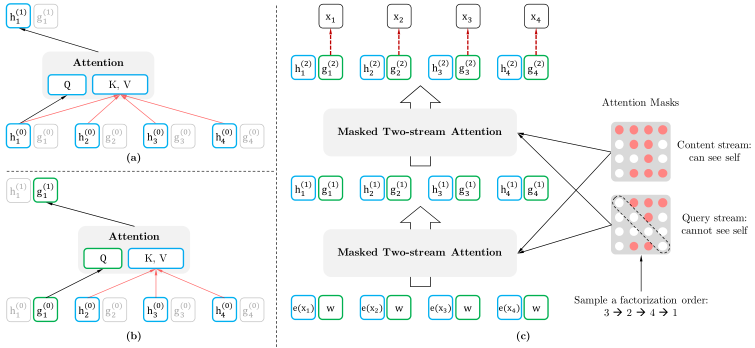}
    \caption{The mask attention of XLNet}
    \label{pipline of text classification}
\end{figure}

\subsection{GPT-2}
 GPT2 is an enhanced version of GPT, and it is based on GPT-2 with the following improvements: GPT-2 collects a larger and more extensive dataset. At the same time, the quality of this dataset is ensured by retaining pages that have high-quality content. Second, GPT-2 increased the number of Transformer stack layers to 48, the hidden layer dimension to 1600, and the number of parameters to 1.5 billion. Third, GPT-2 increased the vocabulary to 50257, the maximum context size from 512 to 1024, and the batch size from 512 to 1024. Self-attention is followed by the addition of a standardization layer; the initialization method of the residual layer is changed, and so on.

\section{Conlusion}
In natural language processing tasks, deep learning-based text classification is a very important research direction. With the development of the Internet and smart phones, the accurate classification of text and analysis of its content has become the frontier in the field of natural language processing. The rapid progress of deep learning has largely replaced the traditional machine learning methods. In this paper, we first introduce feature extraction methods for text classification, such as word2vec, and GloVE, which is a method of mutual inference between central words and contextual words. After that, we briefly introduce some feature reduction methods applied to text classification. We focus on deep learning based text classification models. These two-stage pre-training models consist of a pre-training process and a two-stage model that is based on the pre-training process. fine tuned process. Today, these deep learning based pre-trained models are the mainstay of natural language processing tasks. These models have been fine-tuned and have yielded excellent results in the field of text classification.

\bibliographystyle{unsrtnat}
\bibliography{references}  

\begin{thebibliography}{16}
\providecommand{\natexlab}[1]{#1}
\providecommand{\url}[1]{\texttt{#1}}
\expandafter\ifx\csname urlstyle\endcsname\relax
  \providecommand{\doi}[1]{doi: #1}\else
  \providecommand{\doi}{doi: \begingroup \urlstyle{rm}\Url}\fi

\bibitem[Kowsari et~al.(2019)Kowsari, Meimandi, Heidarysafa, Mendu, Barnes, and
  Brown]{DBLP:journals/corr/abs-1904-08067}
Kamran Kowsari, Kiana~Jafari Meimandi, Mojtaba Heidarysafa, Sanjana Mendu,
  Laura~E. Barnes, and Donald~E. Brown.
\newblock Text classification algorithms: {A} survey.
\newblock \emph{CoRR}, abs/1904.08067, 2019.
\newblock URL \url{http://arxiv.org/abs/1904.08067}.

\bibitem[Abe and Mamitsuka(1998)]{DBLP:conf/icml/AbeM98}
Naoki Abe and Hiroshi Mamitsuka.
\newblock Query learning strategies using boosting and bagging.
\newblock In Jude~W. Shavlik, editor, \emph{Proceedings of the Fifteenth
  International Conference on Machine Learning {(ICML} 1998), Madison,
  Wisconsin, USA, July 24-27, 1998}, pages 1--9. Morgan Kaufmann, 1998.

\bibitem[Chen et~al.(2017)Chen, Xie, Wang, Pradhan, Hong, Bui, Duan, and
  Ma]{CHEN2017147}
Wei Chen, Xiaoshen Xie, Jiale Wang, Biswajeet Pradhan, Haoyuan Hong, Dieu~Tien
  Bui, Zhao Duan, and Jianquan Ma.
\newblock A comparative study of logistic model tree, random forest, and
  classification and regression tree models for spatial prediction of landslide
  susceptibility.
\newblock \emph{CATENA}, 151:\penalty0 147 -- 160, 2017.
\newblock ISSN 0341-8162.
\newblock \doi{https://doi.org/10.1016/j.catena.2016.11.032}.
\newblock URL
  \url{http://www.sciencedirect.com/science/article/pii/S0341816216305136}.

\bibitem[Larson(2010)]{DBLP:journals/jasis/Larson10}
Ray~R. Larson.
\newblock Introduction to information retrieval.
\newblock \emph{J. Assoc. Inf. Sci. Technol.}, 61\penalty0 (4):\penalty0
  852--853, 2010.
\newblock \doi{10.1002/asi.21234}.
\newblock URL \url{https://doi.org/10.1002/asi.21234}.

\bibitem[Mikolov et~al.(2013{\natexlab{a}})Mikolov, Chen, Corrado, and
  Dean]{DBLP:journals/corr/abs-1301-3781}
Tomas Mikolov, Kai Chen, Greg Corrado, and Jeffrey Dean.
\newblock Efficient estimation of word representations in vector space.
\newblock In Yoshua Bengio and Yann LeCun, editors, \emph{1st International
  Conference on Learning Representations, {ICLR} 2013, Scottsdale, Arizona,
  USA, May 2-4, 2013, Workshop Track Proceedings}, 2013{\natexlab{a}}.
\newblock URL \url{http://arxiv.org/abs/1301.3781}.

\bibitem[Pennington et~al.(2014)Pennington, Socher, and
  Manning]{DBLP:conf/emnlp/PenningtonSM14}
Jeffrey Pennington, Richard Socher, and Christopher~D. Manning.
\newblock Glove: Global vectors for word representation.
\newblock In Alessandro Moschitti, Bo~Pang, and Walter Daelemans, editors,
  \emph{Proceedings of the 2014 Conference on Empirical Methods in Natural
  Language Processing, {EMNLP} 2014, October 25-29, 2014, Doha, Qatar, {A}
  meeting of SIGDAT, a Special Interest Group of the {ACL}}, pages 1532--1543.
  {ACL}, 2014.
\newblock \doi{10.3115/v1/d14-1162}.
\newblock URL \url{https://doi.org/10.3115/v1/d14-1162}.

\bibitem[Karamizadeh et~al.(2009)Karamizadeh, Abdullah, Manaf, Zamani, and
  Hooman]{Karamizadeh2009PrincipalCA}
Sasan Karamizadeh, S.~Abdullah, A.~A. Manaf, Mazdak Zamani, and Alireza Hooman.
\newblock Principal component analysis.
\newblock In \emph{Encyclopedia of Biometrics}, 2009.

\bibitem[H{\'e}rault and Ans(1984)]{Hrault1984RseauDN}
J.~H{\'e}rault and B.~Ans.
\newblock R{\'e}seau de neurones {\`a} synapses modifiables: d{\'e}codage de
  messages sensoriels composites par apprentissage non supervis{\'e} et
  permanent.
\newblock 1984.

\bibitem[Hochreiter and Schmidhuber(1997)]{DBLP:journals/neco/HochreiterS97}
Sepp Hochreiter and J{\"{u}}rgen Schmidhuber.
\newblock Long short-term memory.
\newblock \emph{Neural Comput.}, 9\penalty0 (8):\penalty0 1735--1780, 1997.
\newblock \doi{10.1162/neco.1997.9.8.1735}.
\newblock URL \url{https://doi.org/10.1162/neco.1997.9.8.1735}.

\bibitem[Vaswani et~al.(2017)Vaswani, Shazeer, Parmar, Uszkoreit, Jones, Gomez,
  Kaiser, and Polosukhin]{DBLP:conf/nips/VaswaniSPUJGKP17}
Ashish Vaswani, Noam Shazeer, Niki Parmar, Jakob Uszkoreit, Llion Jones,
  Aidan~N. Gomez, Lukasz Kaiser, and Illia Polosukhin.
\newblock Attention is all you need.
\newblock In Isabelle Guyon, Ulrike von Luxburg, Samy Bengio, Hanna~M. Wallach,
  Rob Fergus, S.~V.~N. Vishwanathan, and Roman Garnett, editors, \emph{Advances
  in Neural Information Processing Systems 30: Annual Conference on Neural
  Information Processing Systems 2017, 4-9 December 2017, Long Beach, CA,
  {USA}}, pages 5998--6008, 2017.
\newblock URL \url{http://papers.nips.cc/paper/7181-attention-is-all-you-need}.

\bibitem[Yang et~al.(2019)Yang, Dai, Yang, Carbonell, Salakhutdinov, and
  Le]{DBLP:journals/corr/abs-1906-08237}
Zhilin Yang, Zihang Dai, Yiming Yang, Jaime~G. Carbonell, Ruslan Salakhutdinov,
  and Quoc~V. Le.
\newblock Xlnet: Generalized autoregressive pretraining for language
  understanding.
\newblock \emph{CoRR}, abs/1906.08237, 2019.
\newblock URL \url{http://arxiv.org/abs/1906.08237}.

\bibitem[Devlin et~al.(2018)Devlin, Chang, Lee, and
  Toutanova]{DBLP:journals/corr/abs-1810-04805}
Jacob Devlin, Ming{-}Wei Chang, Kenton Lee, and Kristina Toutanova.
\newblock {BERT:} pre-training of deep bidirectional transformers for language
  understanding.
\newblock \emph{CoRR}, abs/1810.04805, 2018.
\newblock URL \url{http://arxiv.org/abs/1810.04805}.

\bibitem[Mikolov et~al.(2013{\natexlab{b}})Mikolov, Sutskever, Chen, Corrado,
  and Dean]{DBLP:conf/nips/MikolovSCCD13}
Tomas Mikolov, Ilya Sutskever, Kai Chen, Gregory~S. Corrado, and Jeffrey Dean.
\newblock Distributed representations of words and phrases and their
  compositionality.
\newblock In Christopher J.~C. Burges, L{\'{e}}on Bottou, Zoubin Ghahramani,
  and Kilian~Q. Weinberger, editors, \emph{Advances in Neural Information
  Processing Systems 26: 27th Annual Conference on Neural Information
  Processing Systems 2013. Proceedings of a meeting held December 5-8, 2013,
  Lake Tahoe, Nevada, United States}, pages 3111--3119, 2013{\natexlab{b}}.
\newblock URL
  \url{http://papers.nips.cc/paper/5021-distributed-representations-of-words-and-phrases-and-their-compositionality}.

\bibitem[Peters et~al.(2018)Peters, Neumann, Iyyer, Gardner, Clark, Lee, and
  Zettlemoyer]{Peters:2018}
Matthew~E. Peters, Mark Neumann, Mohit Iyyer, Matt Gardner, Christopher Clark,
  Kenton Lee, and Luke Zettlemoyer.
\newblock Deep contextualized word representations.
\newblock In \emph{Proc. of NAACL}, 2018.

\bibitem[Radford(2018)]{Radford2018ImprovingLU}
A.~Radford.
\newblock Improving language understanding by generative pre-training.
\newblock 2018.

\bibitem[Radford et~al.(2019)Radford, Wu, Child, Luan, Amodei, and
  Sutskever]{radford2019language}
Alec Radford, Jeff Wu, Rewon Child, David Luan, Dario Amodei, and Ilya
  Sutskever.
\newblock Language models are unsupervised multitask learners.
\newblock 2019.

\end{thebibliography}






\end{document}